\documentclass[letterpaper, 10 pt, conference, dvipdfmx]{ieeeconf}  

\IEEEoverridecommandlockouts                              

\usepackage{amsmath} 
\usepackage{amssymb}  

\usepackage[dvipdfmx]{graphicx}
\usepackage{url}
\usepackage{multirow}
\usepackage{siunitx}

\newcommand{\figref}[1]{{Fig.~\ref{#1}}}

\newcommand{\bm}[1]{\mbox{\boldmath{$#1$}}}

\title{\LARGE \bf
  Learning Differentiable Reachability Maps for \\Optimization-based Humanoid Motion Generation
}

\author{Masaki Murooka, Iori Kumagai, Mitsuharu Morisawa and Fumio Kanehiro
  \thanks{The authors are with
    CNRS-AIST JRL (Joint Robotics Laboratory), IRL and
    National Institute of Advanced Industrial Science and Technology (AIST),
    1-1-1 Umezono, Tsukuba, Ibaraki 305-8560, Japan.
    {\tt\small m-murooka@aist.go.jp}}%
}

\begin{document}

\maketitle
\thispagestyle{empty}
\pagestyle{empty}

\setlength{\floatsep}{6pt}
\setlength{\textfloatsep}{6pt}
\setlength{\abovecaptionskip}{4pt}
\setlength{\abovedisplayskip}{4pt}
\setlength{\belowdisplayskip}{4pt}

\begin{abstract}
To reduce the computational cost of humanoid motion generation, we introduce a new approach to representing robot kinematic reachability: the \textit{differentiable reachability map}. This map is a scalar-valued function defined in the task space that takes positive values only in regions reachable by the robot's end-effector. A key feature of this representation is that it is continuous and differentiable with respect to task-space coordinates, enabling its direct use as constraints in continuous optimization for humanoid motion planning. We describe a method to learn such differentiable reachability maps from a set of end-effector poses generated using a robot's kinematic model, using either a neural network or a support vector machine as the learning model.  By incorporating the learned reachability map as a constraint, we formulate humanoid motion generation as a continuous optimization problem. We demonstrate that the proposed approach efficiently solves various motion planning problems, including footstep planning, multi-contact motion planning, and loco-manipulation planning for humanoid robots.
\end{abstract}

\section{Introduction} \label{sec:intro}

Motion planning for humanoid robots involves solving complex inverse kinematics (IK) problems under constraints such as joint limits and self-collision avoidance. However, IK is typically formulated as a nonlinear optimization problem that must be solved iteratively, and this computational cost becomes significant when it must be evaluated repeatedly during trajectory generation or whole-body motion planning.

To address this issue, we propose a data-driven method that learns a continuous representation of kinematic reachability, referred to as a differentiable reachability map, from a sample set of end-effector poses. This reachability map provides a scalar-valued function defined over the task space that indicates whether a given pose is reachable, and it is differentiable with respect to task-space coordinates. To construct the map, we adopt data-driven models such as neural networks (NNs) and support vector machines (SVMs), which offer continuous and differentiable representations of reachability. By embedding this differentiable function into the constraints of an optimization problem, we can express reachability conditions in a continuous and differentiable form, avoiding the need for solving IK repeatedly during the optimization.

This paper focuses on the application of the differentiable reachability map to optimization-based motion planning for humanoid robots. We formulate various motion generation problems, such as footstep planning, multi-contact motion planning, and loco-manipulation planning, using reachability constraints expressed by the learned map. The proposed framework enables efficient generation of feasible motions in the task space, while avoiding computationally expensive joint-space optimization.

\section{Related Works and Contributions}

\subsection{Discrete Representations of Kinematic Reachability}

Several studies have been performed to pre-generate kinematically reachable task-space regions (also called reachability maps) to evaluate reachability without solving IK.
The most typical method for pre-generating such task-space regions is to first divide the task space into grids; then, solve IK with each point in the grid as the target; finally, save the set of points where IK was successful~\cite{ReachabilityMap:Zacharias:IROS2007,ReachablePlacement:Vahrenkamp:ICRA2013}.
A library that speeds up the generation process by analytical IK~\cite{OpenRAVE:Rosen:CMU2008} has been used for examining the body structure in robot design~\cite{HRP5P:Kaneko:RAL2019}.
Other generation methods based on the convolution of the movable regions of each joint have also been proposed~\cite{RmapConv:Han:ICRA2021}, but they are also represented in a discretized task space.

Reachability maps have been used to evaluate the feasibility of task-space trajectories and to determine the placement of robots.
In \cite{ReachabilityMap:Zacharias:Humanoids2009,OrientationRmap:Dong:IROS2015,Reuleaux:Makhal:IRC2018}, each point of the discretized target end-effector trajectory is checked for inclusion in the reachability maps to determine the feasibility of the trajectory.
As in these examples, discrete reachability maps are fully available for evaluating a given trajectory in task space. However, they are not suitable for use in trajectory optimization to generate a new trajectory because they cannot be expressed as a continuous objective function or constraints.
Additionally, the task-space dimension increases when considering both the position and orientation, and grid division is problematic in terms of computational cost because the number of grids increases exponentially.

\subsection{Reachability Constraints in Humanoid Motion Planning}

Reachability of end-effectors (e.g., hands and feet) is an important constraint in motion planning.
In graph search-based motion planning, discrete reachability maps are often utilized.
For example, in bipedal footstep planning using A* search~\cite{FootstepPlan:Hornung:Humanoids2012}, the search graph is extended based on a finite set of landing positions of the swing foot relative to the stance foot.
This set of landing positions can be regarded as a kind of discrete reachability map.
In loco-manipulation planning~\cite{Locomanip:Jorgensen:ICRA2020, LocomanipPlan:Murooka:RAL2021}, which is an extension of bipedal footstep planning, pre-generated discrete reachability maps are used to determine the conditions for reaching the manipulated object by hand.

In continuous optimization-based motion planning, the admissible region of end-effectors is often approximated by a box~\cite{PhaseBasedTO:Winkler:RAL2018} or convex polyhedron~\cite{ContactPlanner:Tonneau:TRO2018} in the task space, thereby avoiding the direct treatment of joint configurations~\cite{CentroidalDynamicsPlanning:Dai:Humanoids2014} and reducing the dimension of the search variables.
However, these approximate shapes restrict the full use of the reachable region of end-effectors, and the planning results may be conservative.
Additionally, these shapes are used as constraints on the position of end-effectors, and the constraint expression regarding the orientation of end-effectors is unclear.
This usability of the shapes is acceptable when the contact at the end-effectors is modeled as a point contact, as in the quadruped robot~\cite{PhaseBasedTO:Winkler:RAL2018}, but it is unacceptable when it is modeled as a surface contact, as in the humanoid robot~\cite{CentroidalDynamicsPlanning:Dai:Humanoids2014}.

To overcome these limitations, we propose using reachability maps that are continuous and differentiable with respect to task-space coordinates, and can flexibly represent arbitrarily-shaped feasible regions.
Such representations enable the reachability condition to be directly embedded as constraints in continuous optimization-based motion generation for humanoid robots.

\subsection{Learning-based Representations of Reachability}

Several previous studies have proposed learning continuous representations of reachability from sample data, as an alternative to discretized maps.

One common approach is to model reachability using Gaussian mixture models (GMMs).
In \cite{CatchObject:Kim:TRO2014} and \cite{GMM3DOrientation:Kim:RAS2017}, the reachable regions of a manipulator's end-effector are encoded as probability density functions represented by GMMs in the task space.
Similarly, in \cite{LearningFeasibility:Carpentier:RSS2017}, a GMM is used to represent the feasibility of the humanoid center of mass (CoM), and the resulting scalar field is incorporated into the objective function of motion optimization.

Another group of studies employs NNs to learn task-space feasibility.
In \cite{LearnGraspReachability:Lou:ICRA2020}, an NN is trained to output a reachability score for candidate grasp poses and is used to guide the selection of grasps in sampling-based planning.
In \cite{LearningReachability:Kim:ICRA2021}, one NN is trained to learn a reachability classifier and another to approximate an inverse reachability map from task space to joint space.
In \cite{IKFlow:Ames:RAL2022}, an NN is trained to generate multiple joint configurations that achieve a given end-effector pose.
In \cite{NNFK:Cursi:RAL2022}, forward kinematics and its Jacobian are simultaneously approximated using an NN, with a focus on ensuring the continuity of differential mappings.

These prior works demonstrate the utility of GMMs and NNs in reachability-related tasks such as feasibility estimation and inverse mapping. Some of these methods rely on density estimation, which assumes that the density of task-space samples correlates with reachability. This assumption may not hold due to the nonlinear mapping from joint space to task space~\cite{ReachabilityMap:Zacharias:IROS2007,Reuleaux:Makhal:IRC2018} as will be discussed in Section~\ref{sec:discussion}. To address this, we formulate reachability learning as a binary classification problem and train differentiable models using either NNs or SVMs. The resulting scalar fields are continuous with respect to task-space coordinates and are directly used as constraints in continuous optimization for humanoid motion generation.

\subsection{Contributions of this Paper}

The contributions of this paper are twofold:
\begin{itemize}
  \item We propose a data-driven approach to learning differentiable reachability maps using NNs and SVMs, which can represent arbitrarily shaped feasible regions in task space.
  \item We apply the learned reachability maps as constraints in optimization-based humanoid motion generation, enabling efficient planning of various humanoid motions.
\end{itemize}

\section{Differentiable Reachability Map} \label{sec:rmap}

\subsection{Formal Definition of Reachability Maps} \label{sec:rmap-definition}

\subsubsection{Binary Reachability Map}

We define a binary reachability map of the robot end-effector as a scalar-valued function that satisfies the following:
\begin{align}
  f_R(\bm{r}) = \left\{ \begin{array}{ll}
    1 & \ \ \mathrm{if \ } \bm{r} \mathrm{\ is \ reachable}\\
    -1 & \ \ \mathrm{otherwise}
  \end{array} \right.
  \label{eq:binary-rmap-definition}
\end{align}
$\bm{r}$ is a point in the task space $\mathrm{\bm{W}_T}$ (e.g., $\mathbb{R}^2$, $\mathrm{SE}(3)$).

In the case where a finite set of reachable points in the task space $\mathcal{L}_r = \{ \bm{r}_1, \bm{r}_2, \cdots, \bm{r}_L \}$ is maintained as a reachability map representation, as proposed in~\cite{ReachabilityMap:Zacharias:IROS2007,ReachablePlacement:Vahrenkamp:ICRA2013}, $f_R(\bm{r})$ in \eqref{eq:binary-rmap-definition} is expressed as follows:
\begin{align}
  f_R(\bm{r}) = \left\{ \begin{array}{ll}
    1 & \ \ \exists \, \bm{\bar{r}} \in \mathcal{L}_r \ \ \mathrm{s.t.} \ \ \bm{r} \in V(\bm{\bar{r}}) \\
    -1 & \ \ \mathrm{otherwise}
  \end{array} \right.
  \label{eq:binary-rmap-from-sample-set}
\end{align}
$V(\bm{r}) \subset \mathrm{\bm{W}_T}$ denotes a neighborhood region around $\bm{r}$ (e.g., a voxel or a sphere), whose size depends on the sampling resolution in the task space.

\subsubsection{Differentiable Reachability Map}

We define a scalar-valued function in the task space $\mathrm{\bm{W}_T}$ as a general reachability map if it satisfies the following:
\begin{align}
  \left\{
  \begin{array}{ll}
    f_R(\bm{r}) \geq 0 & \quad \mathrm{if\ } \bm{r} \mathrm{\ is\ reachable} \\
    f_R(\bm{r}) < 0 & \quad \mathrm{otherwise}
  \end{array}
  \right.
  \label{eq:general-rmap-definition}
\end{align}

The binary reachability map defined in \eqref{eq:binary-rmap-definition} is a special case of this general reachability map.

When $f_R(\bm{r})$ is continuous and differentiable, we refer to it as a differentiable reachability map.
Since $f_R(\bm{r})$ in \eqref{eq:binary-rmap-definition} is discontinuous at the boundary between reachable and unreachable regions, it does not qualify as a differentiable reachability map.

In general, a differentiable reachability map cannot be derived analytically from the robot's kinematic model. Therefore, we propose a method to learn it from a task-space sample set $\mathcal{L}_r$.

\subsection{Learning Reachability Maps from Samples}

We formulate the problem of constructing a differentiable reachability map as a binary classification task in the task space.
Given a dataset $\mathcal{D} = \{ (\bm{x}_i, y_i) \}_{i=1}^L$ where each input $\bm{x}_i \in \mathbb{R}^{M_x}$ represents a task-space point and the label $y_i \in \{-1, 1\}$ indicates whether it is reachable, we aim to learn a scalar-valued function $f_R(\bm{x})$ that satisfies the conditions in \eqref{eq:general-rmap-definition}.
This function should be continuous and differentiable so that it can be used as a constraint in continuous optimization for motion generation.

To this end, we consider two types of classifiers: NNs and SVMs, each offering a differentiable formulation of the reachability map.
The following sections describe their respective implementations.

\subsection{Learning with Neural Networks}

We use an NN to represent the differentiable reachability map as a binary classifier.
The NN takes a point in the task space as input and outputs a real-valued scalar, which we interpret via its sign as indicating reachability, as defined in~\eqref{eq:general-rmap-definition}.

The NN is implemented as a multilayer perceptron (MLP) composed of fully connected layers and nonlinear activation functions such as ReLU.
The final output layer remains linear to yield an unconstrained scalar output.
Let $f_R(\bm{x}; \bm{\theta})$ denote the output of the NN parameterized by $\bm{\theta}$, given input $\bm{x}$.
We train the network using a smooth classification loss function, given by:
\begin{align}
  \mathcal{L}(f_R(\bm{x}_i; \bm{\theta}), y_i) &= \mathrm{softplus}(- y_i \, f_R(\bm{x}_i; \bm{\theta})) \nonumber \\
  &= \log(1 + \exp(- y_i \, f_R(\bm{x}_i; \bm{\theta})))
\end{align}
where $(\bm{x}_i, y_i)$ is a labeled sample from the training dataset.
The resulting scalar field is continuous and differentiable over the task space, making it suitable for use as a constraint in trajectory optimization.
The gradient $\partial f_R(\bm{x}; \bm{\theta}) / \partial \bm{x}$ can be automatically computed via backpropagation.

\subsection{Learning with Support Vector Machines}

\subsubsection{Kernel-based Expression}

We use an SVM with a radial basis function (RBF) kernel to represent the differentiable reachability map.
The decision function of the SVM is given by:
\begin{align}
  f_R(\bm{x}) = \sum_{i = 1}^{L} \alpha_i y_i \, \mathrm{exp}(- \gamma \| \bm{x} - \bm{x}_i \|^2) + b \label{eq:predict}
\end{align}
where $\gamma$ is a kernel bandwidth hyperparameter, and the parameters $\alpha_i$ and $b$ are determined through standard SVM training.

This expression satisfies the condition of the general reachability map defined in \eqref{eq:general-rmap-definition}.
It is also differentiable with respect to $\bm{x}$, and the gradient can be computed analytically:
\begin{align}
  \frac{\partial f_R(\bm{x})}{\partial \bm{x}} = -2 \gamma \sum_{i = 1}^{L} \alpha_i y_i \, \mathrm{exp}(- \gamma \| \bm{x} - \bm{x}_i \|^2) \, (\bm{x} - \bm{x}_i)^{\mathsf{T}} \label{eq:predict-grad}
\end{align}

\subsubsection{One-Class SVM for Reachable-Only Data} \label{sec:one-class-SVM}

In some scenarios, such as when sampling task-space points via forward kinematics (FK) from joint-space samples, the dataset may contain only reachable points.
In such cases, a standard binary SVM cannot be trained directly due to the absence of negative samples.

To address this, we employ one-class SVM~\cite{oneClassSVM:Scholkopf:NIPS2000}, a variant designed for unsupervised anomaly detection.
It treats the origin of the feature space as an implicit negative class and learns a boundary that encloses the support of the positive samples.
Although originally intended for anomaly detection, we demonstrate that this method is also effective for constructing reachability maps from reachable-only data.

As will be shown in Section~\ref{sec:rmap-eval}, reachability maps constructed with regular SVM and one-class SVM exhibit different characteristics and can be chosen based on task requirements.

\subsection{Sample Set Generation} \label{sec:sample-generation}

We employ two sampling strategies to generate datasets for training differentiable reachability maps: a FK-based method and an IK-based method.
The FK-based method samples joint configurations within the motion range of the robot's joints and computes the corresponding end-effector poses via FK. Samples that result in self-collisions are discarded. This method yields only reachable poses and is used exclusively for training one-class SVMs.

The IK-based method, in contrast, generates both reachable and unreachable samples. It randomly samples end-effector poses from the task space and solves IK to determine reachability. A pose is labeled as reachable if a valid IK solution exists that satisfies joint limit and self-collision constraints~\cite{CollisionAvoidance:Faverjon:ICRA1987}; otherwise, it is labeled as unreachable. The resulting labeled dataset is used to train both NNs and regular SVMs.

To address the curse of dimensionality in high-dimensional task spaces, we use random sampling rather than grid sampling. Random sampling allows for more scalable data generation and flexible early termination.
Moreover, when a motion planning task involves only a lower-dimensional subspace of the task space, it is more efficient to sample directly in that subspace using the IK-based method. For example, in the footstep planning task described in Section~\ref{sec:footstep-planning}, the foot's reachable space is originally defined over $\mathrm{SE}(3)$, but sampling is restricted to the $\mathrm{SE}(2)$ subspace for efficiency.

\section{Numerical Evaluation} \label{sec:rmap-eval}

\subsection{Comparison with Analytical Reachability}

\begin{figure}[tpb]
  \begin{center}
    \includegraphics[width=0.95\columnwidth]{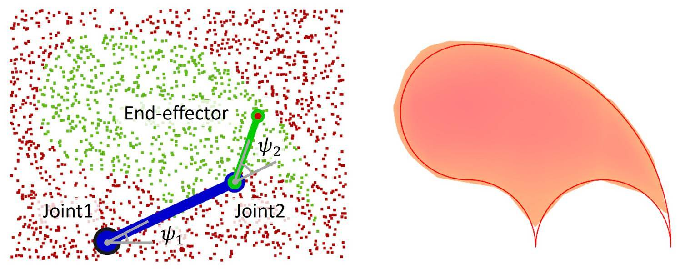}\\
    \begin{minipage}{0.48\columnwidth}
      \begin{center} \footnotesize (A) Sample set \end{center}
    \end{minipage}
    \begin{minipage}{0.48\columnwidth}
      \begin{center} \footnotesize (B) NN map \end{center}
    \end{minipage}\\
    \vspace{1.5mm}
    \includegraphics[width=0.95\columnwidth]{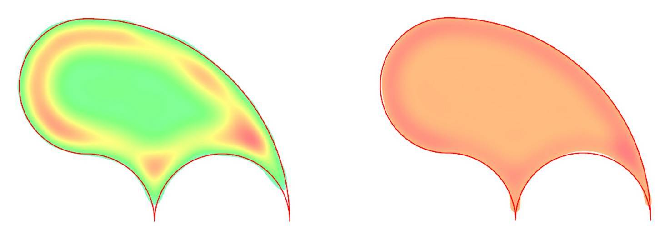}\\
    \begin{minipage}{0.48\columnwidth}
      \begin{center} \footnotesize (C) Regular SVM map \end{center}
    \end{minipage}
    \begin{minipage}{0.48\columnwidth}
      \begin{center} \footnotesize (D) One-class SVM map \end{center}
    \end{minipage}\\
    \vspace{1.5mm}
    \includegraphics[width=1.0\columnwidth]{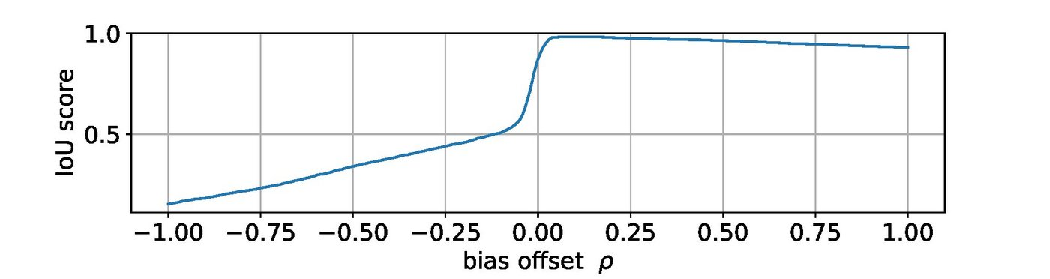}\\
    \vspace{-0.5mm}
    \begin{minipage}{1.0\columnwidth}
      \begin{center} \footnotesize (E) Classification IoU as a function of offset \end{center}
    \end{minipage}
    \caption{Reachability determination for a planar 2-DoF manipulator.
      \newline \footnotesize
      (A) Samples in $\mathbb{R}^2$ space generated by the IK-based method.
      The joint angle limits are $\psi_1 \in [0, \frac{\pi}{2}]$ and $\psi_2 \in [0, \pi]$.
      (B-D) Heatmaps of reachability maps learned by the NN (B), regular SVM (C), and one-class SVM (D).
      The regions where the function values are positive (i.e., predicted reachable) are shown.
      The red curve indicates the analytically derived reachable boundary.
      (E) Classification IoU as a function of the offset $\rho$ added to the SVM decision function. A small positive offset improves the IoU, particularly for the one-class SVM.
    }
    \vspace{-2mm}
    \label{fig:eval1}
  \end{center}
\end{figure}

We evaluated the learned reachability maps using a planar 2-DoF manipulator whose analytical reachable region is known.
As illustrated in Fig.~\ref{fig:eval1}~(A), we generated 10,000 samples using both FK-based and IK-based methods.
The sample generation time was $3.8\,\mathrm{ms}$ for the FK-based method and $8.0\,\mathrm{s}$ for the IK-based method.

We trained both NNs and SVMs using the generated sample sets.
The NN and regular SVM were trained with the IK-based dataset, which contains both reachable and unreachable samples, while the one-class SVM was trained with the FK-based dataset consisting only of reachable samples.
The NN was implemented as an MLP with layer sizes 64, 32, and 1.
The RBF kernel bandwidth for both types of SVMs was set to 30.
Training times were $9.9\,\mathrm{s}$ for the NN, $0.36\,\mathrm{s}$ for the regular SVM, and $0.30\,\mathrm{s}$ for the one-class SVM.

The learned maps are shown in Fig.~\ref{fig:eval1}~(B), (C), and (D).
Each method produces a reachable region that closely matches the analytical ground truth.
The intersection over union (IoU) on a test dataset was 0.971 for the NN, 0.982 for the regular SVM, and 0.983 for the one-class SVM.
The average inference time per sample was approximately $40\,\mathrm{\mu s}$ for the NN and $10\,\mathrm{\mu s}$ for both SVMs.

In these evaluations, a constant offset $\rho$ was added to the SVM decision function~\eqref{eq:predict} to improve classification accuracy, particularly for the one-class SVM.
As shown in Fig.~\ref{fig:eval1}~(E), a small positive offset helps better align the decision boundary.
We used $\rho = 0.1$, determined empirically.
For the NN, no such offset was required, as the classification result was robust to the decision threshold.

\subsection{Comparison with Baseline Classifiers}

We compared our proposed methods (NN, regular SVM (R-SVM), and one-class SVM (OC-SVM)) with baseline classifiers across task spaces of varying dimensionality: $\mathbb{R}^2$, $\mathrm{SE}(2)$, $\mathbb{R}^3$, and $\mathrm{SE}(3)$.
All datasets were generated using the kinematic model of a Universal Robots UR10 manipulator, and performance was evaluated on a held-out test set.

Table~\ref{tab:eval1} presents the IoU scores [\%] for each method, along with per-sample inference time [$\mathrm{ms}$] shown in parentheses.
Unlike our methods, baseline approaches such as $k$-NN ($k$-nearest neighbors) and OC-NN (one-class nearest neighbors~\cite{OCNN:Khan:TKDE2018}) do not yield a scalar-valued function over the task space, and therefore cannot be used as constraints in continuous trajectory optimization.
A convex-hull-based classifier is also included as a simple geometric baseline for low-dimensional spaces.

Overall, our methods match or outperform the baselines in terms of classification accuracy and inference efficiency.
Notably, the NN model outperforms SVMs in the $\mathrm{SE}(3)$ task space in terms of accuracy, inference time, and training cost.
Despite being trained solely on positive samples, OC-SVM performs comparably to classifiers trained with both positive and negative data across all task spaces.

\renewcommand{\arraystretch}{1.2}
\begin{table}[h]
  \caption{Comparison of reachability classification methods}
  \label{tab:eval1}
  \vspace{-4mm}
  \begin{center}
    \begin{tabular}{l||cccc}
      \hline
      Task space & $\mathbb{R}^2$ & $\mathrm{SE}(2)$ & $\mathbb{R}^3$ & $\mathrm{SE}(3)$ \\ \hline\hline
      NN & 96.9 & 96.2 (0.044) & 96.8 (0.040) & 89.3 (0.062) \\ \hline
      R-SVM & 98.5 & 95.1 (0.13) & 97.6 (0.080) & 77.7 (8.7) \\ \hline
      OC-SVM & 98.2 & 96.7 (0.16) & 84.3 (0.091) & 81.1 (7.9) \\ \hline
      $k$-NN & 97.4 & 93.1 (0.19) & 95.7 (0.32) & 83.4 (2.5) \\ \hline
      OC-NN & 87.6 & 81.3 (0.31) & 84.0 (0.31) & 73.3 (4.8) \\ \hline
      Convex & 82.3 & - & - & - \\ \hline\hline
      $N^{\mathrm{train}}_{\mathrm{sample}}$ & 10,000 & 80,000 & 80,000 & 500,000 \\ \hline
      $T^{\mathrm{train}}_{\mathrm{NN}}$ [$\mathrm{s}$] & 10 $\,\mathrm{s}$ & 65 $\,\mathrm{s}$ & 66 $\,\mathrm{s}$ & 428 $\,\mathrm{s}$ \\ \hline
      $T^{\mathrm{train}}_{\mathrm{SVM}}$ [$\mathrm{s}$] & 0.3 $\,\mathrm{s}$ & 30 $\,\mathrm{s}$ & 20 $\,\mathrm{s}$ & $\num{1.6e4} \,\mathrm{s}$ \\
      \hline
    \end{tabular} \\
    \vspace{2mm}
    \begin{minipage}{1.0\columnwidth}
      \footnotesize{
        IoU scores [\%] on the test sample set. Values in parentheses indicate per-sample inference time [$\mathrm{ms}$].
        ``NN'', ``R-SVM'', and ``OC-SVM'' denote our proposed methods based on neural networks, regular SVM, and one-class SVM, respectively.
        ``$k$-NN'' and ``OC-NN'' refer to baseline methods based on $k$-nearest neighbors and one-class nearest neighbors~\cite{OCNN:Khan:TKDE2018}, respectively.
        ``Convex'' classifies based on inclusion in the convex hull of reachable samples.
        OC-SVM, OC-NN, and Convex use only positive training samples; NN, R-SVM, and $k$-NN use both positive and negative samples.
        $N^{\mathrm{train}}_{\mathrm{sample}}$ is the number of training samples. $T^{\mathrm{train}}_{\mathrm{NN}}$ and $T^{\mathrm{train}}_{\mathrm{SVM}}$ indicate training time for NN and SVM, respectively.
      }
    \end{minipage}
    \vspace{-2mm}
  \end{center}
\end{table}

\section{Motion Generation with Reachability Maps} \label{sec:planning}

\subsection{Problem Formulation with Reachability Constraints} \label{sec:core-formulation}

We formulate a motion generation problem that incorporates the differentiable reachability map $f_R(\bm{x})$ as a constraint. A basic example is:

\begin{align}
  \min_{\bm{r} \in \mathrm{\bm{W}_T}} \ \frac{1}{2} \| \bm{r} - \bm{r}^{\mathrm{trg}} \|^2
  \quad \mathrm{s.t.} \quad
  f_R(\bm{x}(\bm{r})) \geq 0
  \label{eq:global-optimization}
\end{align}

Here, $\bm{r}$ is the end-effector pose in the task space $\mathrm{\bm{W}_T}$ (e.g., $\mathbb{R}^2$, $\mathrm{SE}(3)$), and $\bm{r}^{\mathrm{trg}}$ is a desired target pose. The input vector $\bm{x}(\bm{r})$ transforms the pose into the appropriate input format for the learned model. For position-only task spaces, $\bm{x}(\bm{r}) = \bm{r}$. Orientation treatment is discussed in Section~\ref{sec:orientation-treatment}.

Since $f_R$ is generally nonlinear, we solve~\eqref{eq:global-optimization} using sequential local approximations. At each iteration, a local quadratic program (QP) is constructed:

\begin{align}
  & \min_{\Delta \bm{r} \in \mathrm{\bm{W}_T}} \ \frac{1}{2} \Delta \bm{r}^{\mathsf{T}} \Delta \bm{r} +
  (\bm{\bar{r}} - \bm{r}^{\mathrm{trg}})^{\mathsf{T}} \Delta \bm{r} \label{eq:local-optimization} \\
  & \mathrm{s.t.} \quad
  \left. \frac{\partial f_R(\bm{x})}{\partial \bm{x}} \right|_{\bm{x} = \bm{x}(\bm{\bar{r}})}
  \left. \frac{\partial \bm{x}(\bm{r})}{\partial \bm{r}} \right|_{\bm{r} = \bm{\bar{r}}}
  \Delta \bm{r} \geq - f_R(\bm{x}(\bm{\bar{r}})) \nonumber
\end{align}

Here, $\bm{\bar{r}}$ is the current solution estimate. The constraint is a linearization of $f_R(\bm{x}(\bm{r})) \geq 0$ around $\bm{\bar{r}}$. The solution is updated iteratively by $\bm{\bar{r}} \leftarrow \bm{\bar{r}} + \Delta \bm{r}$.

\subsection{Formulation for Humanoid Motion Generation} \label{sec:planning-formulation}

We formulate two types of optimization problems for humanoid motion generation by extending the basic problem introduced in~\eqref{eq:global-optimization}.

\subsubsection{Simultaneous Reachability}

We define the following as a simultaneous reachability problem, in which multiple end-effector target poses must be reachable from a common baselink pose:
\begin{align}
  & \hspace{-4mm}
  \min_{\bm{r}_0, \cdots, \bm{r}_N} \,
  \frac{1}{2} \sum_{i = 1}^{N} \| \bm{r}_i - \bm{r}_i^{\mathrm{trg}} \|^2 +
  \frac{\lambda}{2} \| \bm{r}_0 - \bm{r}_0^{\mathrm{trg}} \|^2
  \label{eq:global-simultaneous-problem} \\
  & \hspace{4mm} \mathrm{s.t.} \quad
  f_R(\bm{x}_{\mathrm{rel}}(\bm{r}_0, \bm{r}_i)) \geq 0 \quad (i = 1, \ldots, N)
  \nonumber
\end{align}

This type of problem arises in robot placement planning, as discussed in Section~\ref{sec:placement-planning}.
While it is typically used for planning the pose of a fixed-base manipulator or a mobile manipulator, it can also be applied to determining the foot placement of a humanoid robot.

Here, $\bm{r}_0$ and $\bm{r}_i \ (i = 1, \ldots, N)$ represent the poses of the baselink and end-effector, respectively.
$\bm{r}_i^{\mathrm{trg}}$ are the target poses for the end-effector, and $\bm{r}_0^{\mathrm{trg}}$ is the optional target pose for the baselink, used when the baselink is not uniquely constrained.
The scalar $\lambda$ weights this regularization term.
$\bm{x}_{\mathrm{rel}}(\bm{r}_0, \bm{r}_i) \in \mathbb{R}^{M_x}$ denotes the input vector to the reachability map model, representing the relative pose from $\bm{r}_0$ to $\bm{r}_i$.
If the task space includes only positional components, then $\bm{x}_{\mathrm{rel}}(\bm{r}_0, \bm{r}_i) = \bm{r}_i - \bm{r}_0$.
Orientation treatment is discussed in Section~\ref{sec:orientation-treatment}.

The corresponding local QP around a tentative solution $\bm{\bar{r}}_0, \ldots, \bm{\bar{r}}_N$ is given by:
\begin{align}
  \min_{\Delta \bm{\hat{r}}} \quad
  & \frac{1}{2} \Delta \bm{\hat{r}}^{\mathsf{T}} \bm{Q} \Delta \bm{\hat{r}} + \bm{c}^{\mathsf{T}} \Delta \bm{\hat{r}}
  \quad \mathrm{s.t.} \quad
  \bm{A} \Delta \bm{\hat{r}} \geq \bm{b}
  \label{eq:local-simultaneous-problem}
\end{align}
where:
\begin{align*}
  \Delta \bm{\hat{r}} &= \begin{bmatrix}
    \Delta \bm{r}_0^{\mathsf{T}} & \Delta \bm{r}_1^{\mathsf{T}} & \cdots & \Delta \bm{r}_N^{\mathsf{T}}
  \end{bmatrix}^{\mathsf{T}}, \\
  \bm{Q} &= \mathrm{diag}(\lambda \bm{I}_{M_r}, \bm{I}_{M_r}, \ldots, \bm{I}_{M_r}), \\
  \bm{c} &= \begin{bmatrix}
    \lambda (\bm{\bar{r}}_0 - \bm{r}_0^{\mathrm{trg}}) \\
    \bm{\bar{r}}_1 - \bm{r}_1^{\mathrm{trg}} \\
    \vdots \\
    \bm{\bar{r}}_N - \bm{r}_N^{\mathrm{trg}}
  \end{bmatrix}, \\
  \bm{A} &= \begin{bmatrix}
    \frac{\partial f_R}{\partial \bm{x}} \frac{\partial \bm{x}_{\mathrm{rel}}}{\partial \bm{r}_0} &
    \frac{\partial f_R}{\partial \bm{x}} \frac{\partial \bm{x}_{\mathrm{rel}}}{\partial \bm{r}_1} & & \\
    \vdots & & \ddots & \\
    \frac{\partial f_R}{\partial \bm{x}} \frac{\partial \bm{x}_{\mathrm{rel}}}{\partial \bm{r}_0} & & &
    \frac{\partial f_R}{\partial \bm{x}} \frac{\partial \bm{x}_{\mathrm{rel}}}{\partial \bm{r}_N}
  \end{bmatrix}, \\
  \bm{b} &= - \begin{bmatrix}
    f_R(\bm{x}_{\mathrm{rel}}(\bm{\bar{r}}_0, \bm{\bar{r}}_1)) \\
    \vdots \\
    f_R(\bm{x}_{\mathrm{rel}}(\bm{\bar{r}}_0, \bm{\bar{r}}_N))
  \end{bmatrix}.
\end{align*}
Here, $\bm{I}_{M_r}$ is the identity matrix of dimension $M_r$, the size of each pose vector $\bm{r}_i$.

\subsubsection{Sequential Reachability}

We define the following as a sequential reachability problem, where each end-effector pose must be reachable from its immediate predecessor:
\begin{align}
  & \hspace{-4mm}
  \min_{\bm{r}_0, \ldots, \bm{r}_N} \,
  \frac{1}{2} \sum_{i \in \{0, N\}} \| \bm{r}_i - \bm{r}_i^{\mathrm{trg}} \|^2 +
  \frac{\lambda}{2} \sum_{i = 0}^{N-1} \| \bm{r}_{i+1} - \bm{r}_i \|^2
  \label{eq:global-sequential-problem} \\
  & \hspace{4mm} \mathrm{s.t.} \quad
  f_R(\bm{x}_{\mathrm{rel}}(\bm{r}_i, \bm{r}_{i+1})) \geq 0 \quad (i = 0, \ldots, N-1)
  \nonumber
\end{align}

This problem formulation is used in footstep planning, as introduced in Section~\ref{sec:footstep-planning}, and also extended to multi-contact and loco-manipulation motion planning in Sections~\ref{sec:multicontact-planning} and \ref{sec:locomanip-planning}, respectively.

Here, $\bm{r}_0, \ldots, \bm{r}_N$ represent the sequence of end-effector poses over time.
The target poses $\bm{r}_0^{\mathrm{trg}}$ and $\bm{r}_N^{\mathrm{trg}}$ are imposed at the start and end of the sequence, and the intermediate poses are regularized to encourage smooth transitions.
The scalar $\lambda$ determines the weight of this regularization.
$\bm{x}_{\mathrm{rel}}(\bm{r}_i, \bm{r}_{i+1})$ denotes the input to the reachability map model, representing the relative pose from $\bm{r}_i$ to $\bm{r}_{i+1}$.

The corresponding local QP around a tentative solution $\bm{\bar{r}}_0, \ldots, \bm{\bar{r}}_N$ shares the same structure as~\eqref{eq:local-simultaneous-problem}, with the following coefficient definitions:
\begin{align}
  \bm{Q} &= \begin{bmatrix}
    \bm{I}_{M_r} & & \\
    & \Large{\bm{O}} & \\
    & & \bm{I}_{M_r}
  \end{bmatrix} +
  \lambda \, \bm{Q}_{\mathrm{adj}},
  \label{eq:local-sequential-problem} \\
  \bm{c} &= \begin{bmatrix}
    \bm{\bar{r}}_0 - \bm{r}_0^{\mathrm{trg}} \\
    \Large{\bm{0}} \\
    \bm{\bar{r}}_N - \bm{r}_N^{\mathrm{trg}}
  \end{bmatrix} +
  \lambda \, \bm{Q}_{\mathrm{adj}}
  \begin{bmatrix}
    \bm{\bar{r}}_0 \\
    \vdots \\
    \bm{\bar{r}}_N
  \end{bmatrix},
  \nonumber \\
  \bm{Q}_{\mathrm{adj}} &= \begin{bmatrix}
    \bm{I}_{M_r} & -\bm{I}_{M_r} & & & \\
    -\bm{I}_{M_r} & 2 \bm{I}_{M_r} & \ddots & & \\
    & -\bm{I}_{M_r} & \ddots & -\bm{I}_{M_r} & \\
    & & \ddots & 2 \bm{I}_{M_r} & -\bm{I}_{M_r} \\
    & & & -\bm{I}_{M_r} & \bm{I}_{M_r}
  \end{bmatrix},
  \nonumber \\
  \bm{A} &= \begin{bmatrix}
    \frac{\partial f_R}{\partial \bm{x}} \frac{\partial \bm{x}_{\mathrm{rel}}}{\partial \bm{r}_0} &
    \frac{\partial f_R}{\partial \bm{x}} \frac{\partial \bm{x}_{\mathrm{rel}}}{\partial \bm{r}_1} & & \\
    & \ddots & \ddots & \\
    & &
    \frac{\partial f_R}{\partial \bm{x}} \frac{\partial \bm{x}_{\mathrm{rel}}}{\partial \bm{r}_{N-1}} &
    \frac{\partial f_R}{\partial \bm{x}} \frac{\partial \bm{x}_{\mathrm{rel}}}{\partial \bm{r}_N}
  \end{bmatrix},
  \nonumber \\
  \bm{b} &= - \begin{bmatrix}
    f_R(\bm{x}_{\mathrm{rel}}(\bm{\bar{r}}_0, \bm{\bar{r}}_1)) \\
    \vdots \\
    f_R(\bm{x}_{\mathrm{rel}}(\bm{\bar{r}}_{N-1}, \bm{\bar{r}}_N))
  \end{bmatrix}.
  \nonumber
\end{align}

\subsection{Treatment of Orientation in Reachability Maps} \label{sec:orientation-treatment}

Orientations in 2D and 3D belong to non-Euclidean spaces and must be handled with care when constructing reachability maps.
Following the approach in~\cite{CatchObject:Kim:TRO2014}, we represent orientations by flattening the elements of the corresponding rotation matrix into a vector. This representation avoids singularities and provides a distance metric that aligns well with rotational similarity, which is advantageous compared to Euler angles or unit quaternions.
For relative poses, the input vector $\bm{x}_{\mathrm{rel}}(\bm{r}_0, \bm{r}_1)$ is constructed in the same manner, using the rotation matrix of the relative orientation between $\bm{r}_0$ and $\bm{r}_1$.

\section{Applications to Humanoid Motion Planning} \label{sec:planning-application}

We present four examples of applying the proposed differentiable reachability maps to humanoid motion planning problems formulated as continuous optimization.
The reachability maps may be modeled using either NNs or SVMs, and both are compatible with the presented formulations.
All planned motions are visualized in the accompanying video.

\subsection{Robot Placement Planning} \label{sec:placement-planning}

Placement planning based on reachability maps is a general strategy that can also be extended to humanoid robots.
To illustrate its core formulation, we first consider a fixed-base manipulator as a simplified case.
This task is modeled as a simultaneous reachability problem~\eqref{eq:global-simultaneous-problem}.

\begin{figure}[tpb]
  \begin{center}
    \includegraphics[width=0.85\columnwidth]{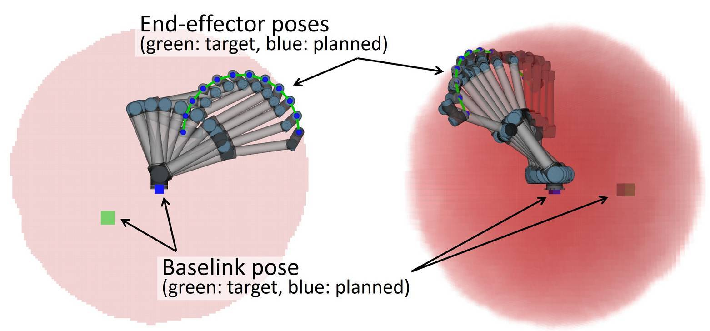}\\
    \vspace{0.5mm}
    \begin{minipage}{0.48\columnwidth}
      \begin{center} \footnotesize (A) Planning in $\mathbb{R}^2$ space \end{center}
    \end{minipage}
    \begin{minipage}{0.46\columnwidth}
      \begin{center} \footnotesize (B) Planning in $\mathbb{R}^3$ space \end{center}
    \end{minipage}
    \vspace{-1mm}
    \caption{Robot placement planning using a reachability map.
      \newline \footnotesize
      Red shading indicates regions where the reachability map evaluates to a positive value (i.e., reachable).
      The baselink target is given lower priority than the end-effector trajectory by assigning a smaller weight in the optimization.
    }
    \vspace{-4mm}
    \label{fig:placement-planning}
  \end{center}
\end{figure}

As shown in Fig.~\ref{fig:placement-planning}, we planned the base pose of a Universal Robots UR10 manipulator so that the end-effector can follow a prescribed trajectory.
The optimization considers both the trajectory of the end-effector and the base pose, allowing for soft prioritization via the regularization term.
In task spaces such as $\mathbb{R}^2$, $\mathbb{R}^3$, and $\mathrm{SE}(2)$, the planner successfully finds base poses that enable the manipulator to reach all waypoints.
In contrast, planning in $\mathrm{SE}(3)$ was more challenging due to the sparsity and complexity of feasible configurations, which is discussed further in Section~\ref{sec:discussion}.
Importantly, the planning cost remains independent of the number of joints, since the optimization is performed directly in task space using the reachability map.

\subsection{Bipedal Footstep Planning} \label{sec:footstep-planning}

We applied the sequential reachability formulation~\eqref{eq:global-sequential-problem} to plan footstep sequences for a humanoid robot.
This approach allows the planned footsteps to satisfy reachability constraints derived from the robot's kinematic model.
The task space is $\mathrm{SE}(2)$, where each variable $\bm{r}_i$ represents the pose of the left or right foot depending on the time step.
The function $f_R(\bm{x}_{\mathrm{rel}}(\bm{r}_i, \bm{r}_{i+1}))$ encodes the feasibility of stepping from one foot to the next, based on learned reachability.

\begin{figure}[tpb]
  \begin{center}
    \includegraphics[width=0.95\columnwidth]{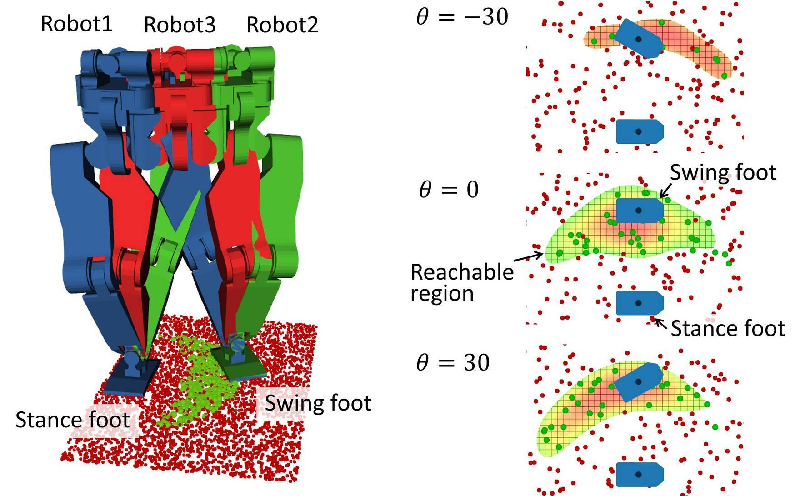} \\
    \vspace{0.5mm}
    \begin{minipage}{0.48\columnwidth}
      \begin{center} \footnotesize (A) Foot pose sampling \end{center}
    \end{minipage}
    \begin{minipage}{0.48\columnwidth}
      \begin{center} \footnotesize (B) Foot reachability map \end{center}
    \end{minipage} \\
    \vspace{2mm}
    \includegraphics[width=0.95\columnwidth]{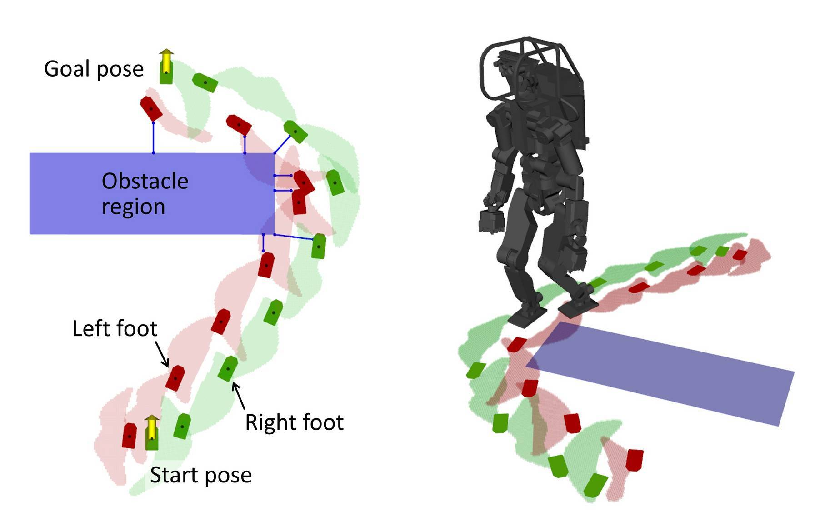} \\
    \vspace{-1mm}
    \begin{minipage}{0.48\columnwidth}
      \begin{center} \footnotesize (C) Footstep planning result \end{center}
    \end{minipage}
    \begin{minipage}{0.48\columnwidth}
      \begin{center} \footnotesize (D) Whole-body motion \end{center}
    \end{minipage}
    \caption{Bipedal footstep planning using reachability maps.
      \newline \footnotesize
      (A) Sampled swing foot poses with respect to stance foot.
      Robots 1, 2, and 3 illustrate cases where the waist link is located on the stance foot, the swing foot, and their midpoint, respectively.
      Green and red points indicate reachable and unreachable samples.
      (B) Heatmaps showing reachability values above zero (i.e., classified as reachable).
      Only samples near the specified yaw angle $\theta$ [deg] are visualized.
      The maps reflect how reachability varies with the orientation of the swing foot relative to the stance foot.
      (C) Footstep planning result.
      The shaded regions indicate the swing foot's reachable area relative to each stance foot pose.
      (D) Whole-body motion generated from the planned footstep sequence and executed in the dynamics simulator.
    }
    \label{fig:footstep-planning}
  \end{center}
\end{figure}

As shown in Fig.~\ref{fig:footstep-planning}~(A), we used the IK-based method to generate sample sets, and trained reachability maps in $\mathrm{SE}(2)$.
Fig.~\ref{fig:footstep-planning}~(B) illustrates the resulting maps, showing how the swing foot's reachable region varies with its orientation relative to the stance foot.
With these maps, our method plans footstep sequences for HRP humanoid (Fig.~\ref{fig:footstep-planning}~(C)) that are feasible under kinematic constraints.
Furthermore, by adding linear inequality constraints for obstacle avoidance~\cite{CollisionAvoidance:Faverjon:ICRA1987}, we can generate footstep sequences that avoid collisions while maintaining feasibility.
Fig.~\ref{fig:footstep-planning}~(D) shows execution results in the dynamics simulator Choreonoid~\cite{Choreonoid:Nakaoka:SII2012}.

\subsection{Multi-contact Motion Planning} \label{sec:multicontact-planning}

The same reachability-based formulation can be extended to multi-contact scenarios, where the robot uses both feet and a hand to make contact.
We planned contact sequences for HRP humanoid walking in a narrow corridor while maintaining balance with a hand on the wall.
The optimization problem extends the sequential reachability formulation~\eqref{eq:global-sequential-problem} to include both foot and hand positions as decision variables.
We assume a fixed sequence of contact transitions, repeating right foot, left foot, and left hand.
The CoM, approximated by the waist link, is assumed to lie on the stance foot during single support, and on the midpoint of both feet during double support.
Reachability constraints are imposed between the CoM and each contacting limb at every phase.

\begin{figure}[tpb]
  \begin{center}
    \includegraphics[width=0.95\columnwidth]{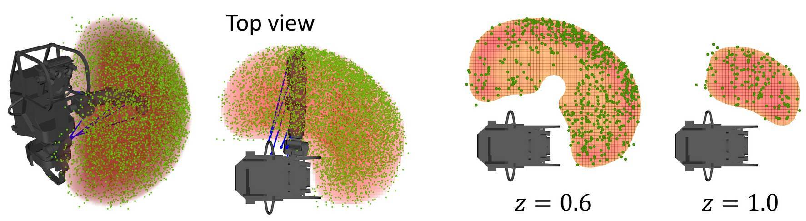} \\
    \vspace{0.5mm}
    \begin{minipage}{0.48\columnwidth}
      \begin{center} \footnotesize (A) Hand position sampling \end{center}
    \end{minipage}
    \begin{minipage}{0.48\columnwidth}
      \begin{center} \footnotesize (B) Hand reachability map \end{center}
    \end{minipage} \\
    \vspace{4mm}
    \includegraphics[width=0.95\columnwidth]{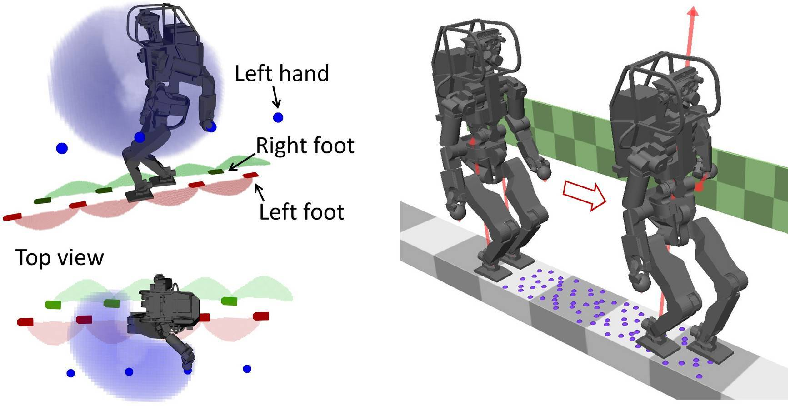} \\
    \vspace{0.5mm}
    \begin{minipage}{0.48\columnwidth}
      \begin{center} \footnotesize (C) Contact planning result\end{center}
    \end{minipage}
    \begin{minipage}{0.48\columnwidth}
      \begin{center} \footnotesize (D) Whole-body motion \end{center}
    \end{minipage}
    \caption{Multi-contact motion planning using reachability maps.
      \newline \footnotesize
      (A) Reachable hand poses sampled using FK. Only reachable samples are shown, as FK-based sampling does not produce unreachable configurations.
      (B) Heatmap of hand reachability values above zero (i.e., classified as reachable) at a specified height $z$ [m].
      (C) Contact sequence planned based on reachability constraints. Shaded areas indicate feasible contact regions for feet and hand.
      (D) Whole-body motion execution in dynamics simulation using the planned contact sequence.
    }
    \label{fig:multicontact-planning}
  \end{center}
\end{figure}

The reachability map for the hand was trained in $\mathbb{R}^3$ assuming point contact, using FK-based samples (Fig.~\ref{fig:multicontact-planning}~(A) and~(B)).
The planned contact sequence satisfies all reachability conditions, and the resulting motion is executed using a multi-contact controller~\cite{Motion6DoF:Murooka:RAL2022}, as shown in Fig.~\ref{fig:multicontact-planning}~(C) and~(D).

\subsection{Loco-manipulation Planning} \label{sec:locomanip-planning}

\begin{figure}[tpb]
  \begin{center}
    \includegraphics[width=0.95\columnwidth]{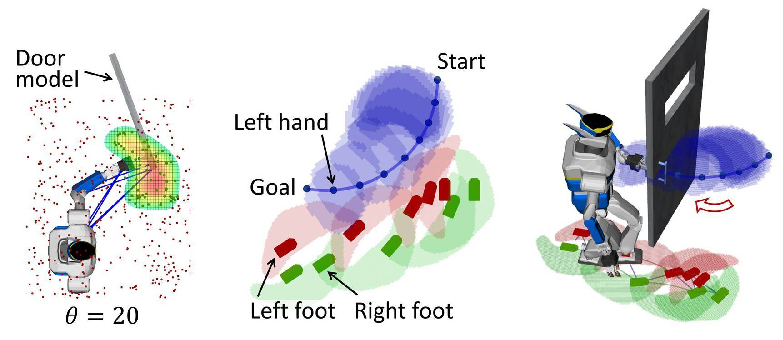} \\
    \vspace{0.5mm}
    \begin{minipage}{0.27\columnwidth}
      \begin{center} \footnotesize (A) Hand sampling \end{center}
    \end{minipage}
    \begin{minipage}{0.35\columnwidth}
      \begin{center} \footnotesize (B) Planning result \end{center}
    \end{minipage}
    \begin{minipage}{0.35\columnwidth}
      \begin{center} \footnotesize (C) Whole-body motion \end{center}
    \end{minipage}
    \caption{Loco-manipulation planning using reachability maps.
      \newline \footnotesize
      (A) Hand samples for reachability map generation, shown with respect to the waist link.
      (B) Planned footstep and hand contact sequence enabling the hand to follow the specified door-opening trajectory.
      (C) Whole-body motion generated from the planned contact sequence.
    }
    \label{fig:locomanip-planning}
  \end{center}
\end{figure}

Loco-manipulation involves walking while simultaneously moving the hand along a specified trajectory.
We formulate this task as an extension of the sequential reachability formulation~\eqref{eq:global-sequential-problem}, incorporating hand reachability constraints alongside footstep planning.
Unlike in multi-contact planning, the hand pose is not directly optimized.
Instead, we introduce a scalar parameter $s$ that defines a predefined hand trajectory and optimize this parameter within the planning framework.
As shown in Fig.~\ref{fig:locomanip-planning}, we plan a motion for HRP humanoid to open a door while maintaining grasp on the doorknob throughout the walking sequence.
The hand trajectory is parameterized by the door opening angle $s$ as $\bm{r}_h(s) = [l_h \!\cos(s) \ l_h \!\sin(s) \ s]^{\mathsf{T}} \in \mathrm{SE}(2)$, where $l_h$ denotes the radius of the doorknob's arc path.

Using reachability maps for both hands and feet trained in $\mathrm{SE}(2)$, the method plans a contact sequence that satisfies the reachability conditions required to follow the trajectory.
Even without a predefined goal pose for the feet, the planner finds a footstep sequence that enables the hand to stay on the trajectory.
The resulting whole-body motion is generated using the method proposed in~\cite{LocomanipPlan:Murooka:RAL2021}.

\section{Discussion} \label{sec:discussion}

\subsubsection{Classification over Density for Reachability Maps}

\begin{figure}[tpb]
  \begin{center}
    \includegraphics[width=0.98\columnwidth, clip, bb=70 20 650 620]{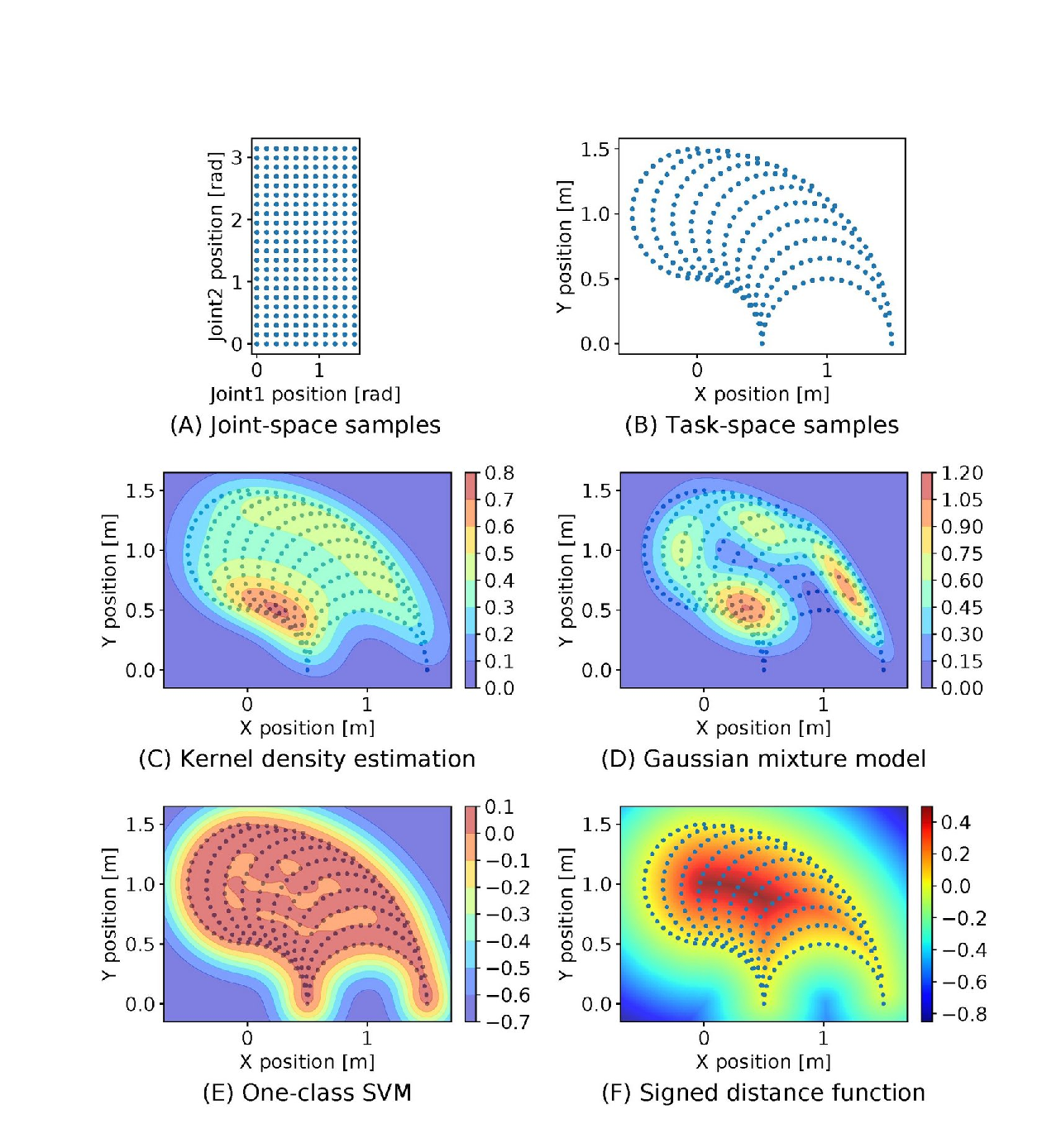}
    \caption{Comparison of scalar-valued functions trained from uniform joint-space samples for a 2-DoF manipulator.
      \newline \footnotesize
      (A) Uniformly sampled configurations in joint space.
      (B) Corresponding distribution in task space, showing non-uniform density.
      (C) Kernel density estimate.
      (D) Gaussian mixture model.
      (E) SVM decision function used as a reachability map.
      (F) Signed distance function to the reachable region.
    }
    \vspace{-4mm}
    \label{fig:compare-density2d}
  \end{center}
\end{figure}

\figref{fig:compare-density2d} illustrates scalar-valued functions learned from uniformly sampled joint configurations of the 2-DoF manipulator shown in Fig.~\ref{fig:eval1}.
As seen in \figref{fig:compare-density2d}~(A) and (B), uniform sampling in joint space leads to non-uniform sample distributions in task space.
As a result, the values obtained through kernel density estimation or Gaussian mixture models (commonly used in previous work~\cite{LearningFeasibility:Carpentier:RSS2017,GMM3DOrientation:Kim:RAS2017}) do not correspond to the margin from the boundary of the reachable region, as shown in \figref{fig:compare-density2d}~(C) and (D).

From this observation, we treat reachability estimation as a classification problem rather than a density estimation task.
The learned function is thus used as an inequality constraint in optimization, not as an objective function.

\subsubsection{Gradient Behavior of the Reachability Map}

\figref{fig:compare-density2d}~(E) shows the decision function of the SVM-based reachability map proposed in this work.
The function remains nearly constant, that is, with near-zero gradient, outside the reachable region far from the boundary.
In the optimization framework described in Section~\ref{sec:planning}, which relies on iteratively solving a local QP with linearly approximated constraints, such zero gradients make it difficult to return to feasible regions once a tentative solution becomes infeasible.

To mitigate this issue, we impose upper and lower bounds on the update step $\Delta \bm{r}$ in the local QP~\eqref{eq:local-optimization}, preventing the optimizer from stepping too far outside the reachable region. This corresponds to a trust region, within which local approximations remain valid~\cite{NumericalOptimization:Nocedal:Springer2006}.

For comparison, \figref{fig:compare-density2d}~(F) shows a signed distance function of the reachable region.
It satisfies the property \eqref{eq:general-rmap-definition} and provides a useful gradient that always points toward the closest point on the boundary.
While this behavior is more optimization-friendly, constructing such functions efficiently in high-dimensional spaces such as $\mathrm{SE}(3)$ remains an open challenge.

\subsubsection{Computation in $\mathrm{SE}(3)$ Space}

As shown in Table~\ref{tab:eval1}, classification accuracy in $\mathrm{SE}(3)$ space is lower than in lower-dimensional task spaces.
This drop in performance is most noticeable near the boundary of the reachable region, and directly affects the feasibility of the planned motion.
To improve performance, promising directions include stricter handling of 3D orientation representations~\cite{GMM3DOrientation:Kim:RAS2017} and adaptive sampling focused near reachability boundaries~\cite{DepthApprox:Pan:TOG2013}.

\section{Conclusion} \label{sec:conclusion}

We proposed a method for learning robot kinematic reachability as a differentiable scalar-valued function, implemented using either NNs or SVMs.
We also presented a formulation for incorporating the learned reachability map into continuous optimization and demonstrated its application to a range of humanoid motion planning problems.
This approach provides a data-driven representation of kinematics that facilitates efficient optimization-based planning for high-degree-of-freedom robots.



\bibliographystyle{IEEEtran}
\bibliography{main.bib}

\begin{thebibliography}{10}
\providecommand{\url}[1]{#1}
\csname url@rmstyle\endcsname
\providecommand{\newblock}{\relax}
\providecommand{\bibinfo}[2]{#2}
\providecommand\BIBentrySTDinterwordspacing{\spaceskip=0pt\relax}
\providecommand\BIBentryALTinterwordstretchfactor{4}
\providecommand\BIBentryALTinterwordspacing{\spaceskip=\fontdimen2\font plus
\BIBentryALTinterwordstretchfactor\fontdimen3\font minus
  \fontdimen4\font\relax}
\providecommand\BIBforeignlanguage[2]{{%
\expandafter\ifx\csname l@#1\endcsname\relax
\typeout{** WARNING: IEEEtran.bst: No hyphenation pattern has been}%
\typeout{** loaded for the language `#1'. Using the pattern for}%
\typeout{** the default language instead.}%
\else
\language=\csname l@#1\endcsname
\fi
#2}}

\bibitem{ReachabilityMap:Zacharias:IROS2007}
F.~Zacharias, C.~Borst, and G.~Hirzinger, ``Capturing robot workspace
  structure: representing robot capabilities,'' in \emph{IEEE/RSJ International
  Conference on Intelligent Robots and Systems}, 2007, pp. 3229--3236.

\bibitem{ReachablePlacement:Vahrenkamp:ICRA2013}
N.~Vahrenkamp, T.~Asfour, and R.~Dillmann, ``Robot placement based on
  reachability inversion,'' in \emph{IEEE International Conference on Robotics
  and Automation}, 2013, pp. 1970--1975.

\bibitem{OpenRAVE:Rosen:CMU2008}
R.~Diankov and J.~Kuffner, ``Open{RAVE}: A planning architecture for autonomous
  robotics,'' \emph{Robotics Institute, Pittsburgh, PA, Tech. Rep.
  CMU-RI-TR-08-34}, vol.~79, 2008.

\bibitem{HRP5P:Kaneko:RAL2019}
K.~{Kaneko}, H.~{Kaminaga}, T.~{Sakaguchi}, S.~{Kajita}, M.~{Morisawa},
  I.~{Kumagai}, and F.~{Kanehiro}, ``{Humanoid Robot HRP-5P}: An electrically
  actuated humanoid robot with high-power and wide-range joints,'' \emph{IEEE
  Robotics and Automation Letters}, vol.~4, no.~2, pp. 1431--1438, 2019.

\bibitem{RmapConv:Han:ICRA2021}
Y.~Han, J.~Pan, M.~Xia, L.~Zeng, and Y.-J. Liu, ``Efficient {SE}(3)
  reachability map generation via interplanar integration of intra-planar
  convolutions,'' in \emph{IEEE International Conference on Robotics and
  Automation}, 2021, pp. 1854--1860.

\bibitem{ReachabilityMap:Zacharias:Humanoids2009}
F.~{Zacharias}, W.~{Sepp}, C.~{Borst}, and G.~{Hirzinger}, ``Using a model of
  the reachable workspace to position mobile manipulators for 3-{D}
  trajectories,'' in \emph{IEEE-RAS International Conference on Humanoid
  Robots}, 2009, pp. 55--61.

\bibitem{OrientationRmap:Dong:IROS2015}
J.~Dong and J.~C. Trinkle, ``Orientation-based reachability map for robot base
  placement,'' in \emph{IEEE/RSJ International Conference on Intelligent Robots
  and Systems}, 2015, pp. 1488--1493.

\bibitem{Reuleaux:Makhal:IRC2018}
A.~Makhal and A.~K. Goins, ``Reuleaux: Robot base placement by reachability
  analysis,'' in \emph{IEEE International Conference on Robotic Computing},
  2018, pp. 137--142.

\bibitem{FootstepPlan:Hornung:Humanoids2012}
A.~{Hornung}, A.~{Dornbush}, M.~{Likhachev}, and M.~{Bennewitz}, ``Anytime
  search-based footstep planning with suboptimality bounds,'' in \emph{IEEE-RAS
  International Conference on Humanoid Robots}, 2012, pp. 674--679.

\bibitem{Locomanip:Jorgensen:ICRA2020}
S.~J. Jorgensen, M.~Vedantam, R.~Gupta, H.~Cappel, and L.~Sentis, ``Finding
  locomanipulation plans quickly in the locomotion constrained manifold,'' in
  \emph{IEEE International Conference on Robotics and Automation}, 2020, pp.
  6611--6617.

\bibitem{LocomanipPlan:Murooka:RAL2021}
M.~Murooka, I.~Kumagai, M.~Morisawa, F.~Kanehiro, and A.~Kheddar, ``Humanoid
  loco-manipulation planning based on graph search and reachability maps,''
  \emph{IEEE Robotics and Automation Letters}, vol.~6, no.~2, pp. 1840--1847,
  2021.

\bibitem{PhaseBasedTO:Winkler:RAL2018}
A.~W. Winkler, C.~D. Bellicoso, M.~Hutter, and J.~Buchli, ``Gait and trajectory
  optimization for legged systems through phase-based end-effector
  parameterization,'' \emph{IEEE Robotics and Automation Letters}, vol.~3,
  no.~3, pp. 1560--1567, 2018.

\bibitem{ContactPlanner:Tonneau:TRO2018}
S.~{Tonneau}, A.~{Del Prete}, J.~{Pettré}, C.~{Park}, D.~{Manocha}, and
  N.~{Mansard}, ``An efficient acyclic contact planner for multiped robots,''
  \emph{IEEE Transactions on Robotics}, vol.~34, no.~3, pp. 586--601, 2018.

\bibitem{CentroidalDynamicsPlanning:Dai:Humanoids2014}
H.~Dai, A.~Valenzuela, and R.~Tedrake, ``Whole-body motion planning with
  centroidal dynamics and full kinematics,'' in \emph{IEEE-RAS International
  Conference on Humanoid Robots}, 2014, pp. 295--302.

\bibitem{CatchObject:Kim:TRO2014}
S.~Kim, A.~Shukla, and A.~Billard, ``Catching objects in flight,'' \emph{IEEE
  Transactions on Robotics}, vol.~30, no.~5, pp. 1049--1065, 2014.

\bibitem{GMM3DOrientation:Kim:RAS2017}
S.~Kim, R.~Haschke, and H.~Ritter, ``Gaussian mixture model for 3-{DoF}
  orientations,'' \emph{Robotics and Autonomous Systems}, vol.~87, pp. 28--37,
  2017.

\bibitem{LearningFeasibility:Carpentier:RSS2017}
J.~Carpentier, R.~Budhiraja, and N.~Mansard, ``Learning feasibility constraints
  for multi-contact locomotion of legged robots,'' in \emph{Robotics: Science
  and Systems}, 2017.

\bibitem{LearnGraspReachability:Lou:ICRA2020}
X.~Lou, Y.~Yang, and C.~Choi, ``Learning to generate 6-{DoF} grasp poses with
  reachability awareness,'' in \emph{IEEE International Conference on Robotics
  and Automation}, 2020, pp. 1532--1538.

\bibitem{LearningReachability:Kim:ICRA2021}
S.~Kim and J.~Perez, ``Learning reachable manifold and inverse mapping for a
  redundant robot manipulator,'' in \emph{IEEE International Conference on
  Robotics and Automation}, 2021, pp. 4731--4737.

\bibitem{IKFlow:Ames:RAL2022}
B.~Ames, J.~Morgan, and G.~Konidaris, ``{IKFlow}: Generating diverse inverse
  kinematics solutions,'' \emph{IEEE Robotics and Automation Letters}, vol.~7,
  no.~3, pp. 7177--7184, 2022.

\bibitem{NNFK:Cursi:RAL2022}
F.~Cursi, W.~Bai, W.~Li, E.~M. Yeatman, and P.~Kormushev, ``Augmented neural
  network for full robot kinematic modelling in {SE(3)},'' \emph{IEEE Robotics
  and Automation Letters}, vol.~7, no.~3, pp. 7140--7147, 2022.

\bibitem{oneClassSVM:Scholkopf:NIPS2000}
B.~Sch\"{o}lkopf, R.~Williamson, A.~Smola, J.~Shawe-Taylor, and J.~Platt,
  ``Support vector method for novelty detection,'' in \emph{International
  Conference on Neural Information Processing Systems}, 1999, pp. 582--588.

\bibitem{CollisionAvoidance:Faverjon:ICRA1987}
B.~Faverjon and P.~Tournassoud, ``A local based approach for path planning of
  manipulators with a high number of degrees of freedom,'' in \emph{IEEE
  International Conference on Robotics and Automation}, vol.~4, 1987, pp.
  1152--1159.

\bibitem{OCNN:Khan:TKDE2018}
S.~S. Khan and A.~Ahmad, ``Relationship between variants of one-class nearest
  neighbors and creating their accurate ensembles,'' \emph{IEEE Transactions on
  Knowledge and Data Engineering}, vol.~30, no.~9, pp. 1796--1809, 2018.

\bibitem{Choreonoid:Nakaoka:SII2012}
S.~{Nakaoka}, ``Choreonoid: Extensible virtual robot environment built on an
  integrated gui framework,'' in \emph{IEEE/SICE International Symposium on
  System Integration}, 2012, pp. 79--85.

\bibitem{Motion6DoF:Murooka:RAL2022}
M.~Murooka, M.~Morisawa, and F.~Kanehiro, ``Centroidal trajectory generation
  and stabilization based on preview control for humanoid multi-contact
  motion,'' \emph{IEEE Robotics and Automation Letters}, vol.~7, no.~3, pp.
  8225--8232, 2022.

\bibitem{NumericalOptimization:Nocedal:Springer2006}
J.~Nocedal and S.~J. Wright, \emph{Numerical optimization}.\hskip 1em plus
  0.5em minus 0.4em\relax Springer, 2006.

\bibitem{DepthApprox:Pan:TOG2013}
J.~Pan, X.~Zhang, and D.~Manocha, ``Efficient penetration depth approximation
  using active learning,'' \emph{ACM Transactions on Graphics}, vol.~32, no.~6,
  2013.

\end{thebibliography}

\end{document}